\documentclass[10pt,twocolumn,letterpaper]{article}

\usepackage{cvpr}
\usepackage{times}
\usepackage{epsfig}
\usepackage{graphicx}
\usepackage{amsmath}
\usepackage{amssymb}
\usepackage{tabularx}
\usepackage[utf8]{inputenc}
\usepackage[T1]{fontenc}
\usepackage{textcomp}
\usepackage{gensymb}
\usepackage{multirow}

\allowdisplaybreaks[2]

\usepackage[pagebackref=true,breaklinks=true,letterpaper=true,colorlinks,bookmarks=false]{hyperref}

\cvprfinalcopy 

\def\cvprPaperID{16} 

\ifcvprfinal\fi

\DeclareMathOperator*{\se3}{\mathfrak{se}(3)}

\begin{document}

\title{Geometric Consistency for Self-Supervised End-to-End Visual Odometry}

\author{Ganesh Iyer$^{*1}$, J. Krishna Murthy$^{*2}$\thanks{The first two authors contributed equally to this work. We thank NVIDIA for donating a DGX-1 computer used in this work. This research was enabled in part by support provided by Compute Canada \href{www.computecanada.ca}{www.computecanada.ca}}, Gunshi Gupta$^1$, K. Madhava Krishna$^{1}$, Liam Paull$^{2}$\\
$^1$ International Institute of Information Technology Hyderabad (India)\\
$^2$ Montreal Institute of Learning Algorithms (MILA), Universite de Montreal\\
{\tt\small \{giyer2309, krrish94, gunshigupta9\}@gmail.com}
\and
}

\maketitle


\begin{abstract} 
With the success of deep learning based approaches in tackling challenging problems in computer vision, a wide range of deep architectures have recently been proposed for the task of visual odometry (VO) estimation. Most of these proposed solutions rely on supervision, which requires the acquisition of precise ground-truth camera pose information, collected using expensive motion capture systems or high-precision IMU/GPS sensor rigs. In this work, we propose an unsupervised paradigm for deep visual odometry learning. We show that using a \emph{noisy teacher}, which could be a standard VO pipeline, and by designing a loss term that enforces geometric consistency of the trajectory, we can train \emph{accurate} deep models for VO that do not require ground-truth labels. We leverage geometry as a self-supervisory signal and propose "Composite Transformation Constraints (CTCs)", that automatically generate supervisory signals for training and enforce geometric consistency in the VO estimate. We also present a method of characterizing the uncertainty in VO estimates thus obtained. To evaluate our VO pipeline, we present exhaustive ablation studies that demonstrate the efficacy of end-to-end, self-supervised methodologies to train deep models for monocular VO. We show that leveraging concepts from geometry and incorporating them into the training of a recurrent neural network results in performance competitive to supervised deep VO methods.
\vspace{-0.5cm}
\end{abstract}

\section{Introduction}


Visual odometry (VO) is the process of estimating the ego-motion of a camera solely from a sequence of images it captures. This capability forms the backbone of any system that requires visual localization. Most solutions to the problems of visual odometry estimation and simultaneous localization and mapping (simultaneously estimating camera trajectory and building a representation of the world) rely on the use of feature matching/tracking or geometric methods in combination with keyframe-based optimization or bundle adjustment \cite{SVO,orb2}. One major challenge of such approaches is to design visual features that have good invariance properties and can be reliably associated. In contrast, deep learning methods \emph{learn} feature representations instead of handcrafting them. Consequently, they have been applied to the problem of visual place recognition for SLAM (discovering that we are in a previously visited place) \cite{Lowry_review} as well as VO \cite{vinet,L-VO,undeepvo_icra,demon,Wang_IJRR,deepvo,sfmlearner,sfmlearner}.

\begin{figure}[t]
\begin{center}
   \includegraphics[width=0.7\linewidth]{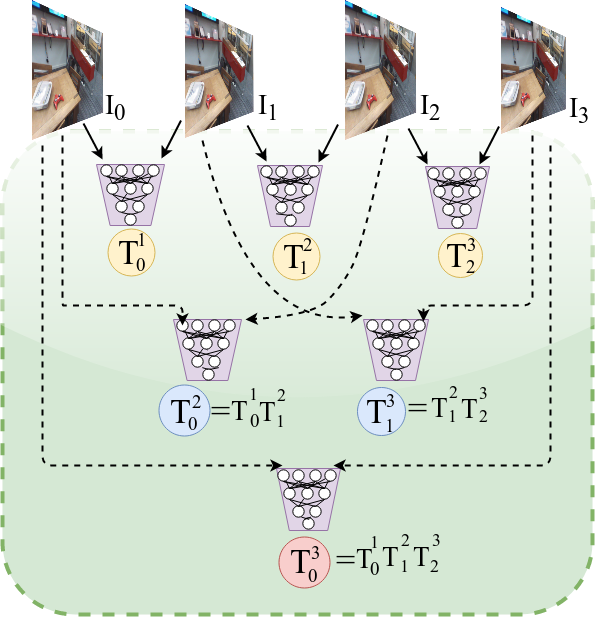}
\end{center}
   \caption{\textbf{System overview}: We leverage the observation that compounded sequences of transformations over short timescales should be equivalent to single transformations over longer timescales. This allows us to create additional constraints, that we refer to as "Composite Transformation Constraints", that can be used as consistency enforcers and aid in training deep architectures for VO without requiring ground-truth labels.}
\label{fig:overview}
\vspace{0.08cm}
\end{figure}

Most learning-based approaches to VO fall into one of the following categories:
\begin{itemize}
    \item Supervised deep VO approaches  assume the availability of ground-truth information in the form of per-frame camera pose in a global frame, usually gathered using a motion-capture system or expensive IMU/GPS sensor rigs \cite{vinet,posenet,dpcnet,demon,Wang_IJRR,deepvo}.
    \item Unsupervised deep VO approaches do not require ground-truth pose information, but leverage 
    some alternate visual information that can assist the learning process, such as depth \cite{gvnn,undeepvo_icra}, stereo images \cite{LRconsistency,sfmlearner}, or optical flow \cite{L-VO}.
\vspace{-0.1cm}
\end{itemize}

Most state-of-the-art deep approaches to VO employ sequence-models, such as long-short term memory (LSTM) units \cite{lstm}, to capture long term dependencies in camera motion \cite{vinet,undeepvo_icra,Wang_IJRR,deepvo}. These models have been shown to correct drift in the estimated trajectory that may have been caused due to incorrect odometry estimates for a few frames in the sequence. However, existing approaches (that do not use depth information) lack tight consistency constraints across time steps. They rely solely on the statefulness of the LSTM model to bring about a weak \emph{smoothing} effect.

We propose an unsupervised training scheme through our proposed model, CTCNet, for the task of learning VO estimation. We tackle the problem in a setting that does not assume the availability of ground-truth odometry data. To this end, we use \emph{noisy} odometry estimates from a conventional VO pipeline (ORB-SLAM \cite{orb2}) to train a recurrent architecture that outputs the relative camera pose transformation between frames. 
To compensate for noisy estimates used in training, we leverage geometry as a self-supervisory signal, and define a set of \emph{Composite Transformation Constraints (CTCs)} across a series of image frames. These constraints arise naturally from the composition law for rigid-body transformations. Estimated transforms over short timescales, when compounded, must equal their counterparts that are computed (independently) over longer timescales. Fig. \ref{fig:overview} shows an example of CTCs applied to an input image sequence comprising four frames. One such constraint here is that compositions of relative transforms between successive frames should equal the transform between the first and the fourth frames. For this to be meaningful however, we require that the longer timescale estimate (i.e., between the first and the fourth frames here) be computed independently. 


In contrast to other works that estimate poses using deep learning \cite{posenet,dpcnet,deepvo}, our network directly regresses to $\se3$ exponential coordinates, and our loss function is formulated as an $L2$-norm over the coordinates. Furthermore, we also describe covariance recovery for VO estimates from our pipeline, using dropout \cite{dropout} to perform approximate Bayesian inference \cite{bayesianPosenet}.

Our experiments on the 7-Scenes \cite{7scenes} dataset demonstrate comparable, and in some cases, better performance compared with supervised methods. We also evaluate several variants of the proposed architecture and demonstrate the flexibility of this training process. To the best of our knowledge, this is the first approach to unsupervised VO estimation that does not require depth prediction as an auxiliary task, as is usually the case \cite{undeepvo_icra,sfm-net,sfmlearner}. 

\section{Related Work}
\vspace{-0.1cm}

\begin{figure*}[!ht]
\begin{center}
    \includegraphics[width=0.9\linewidth]{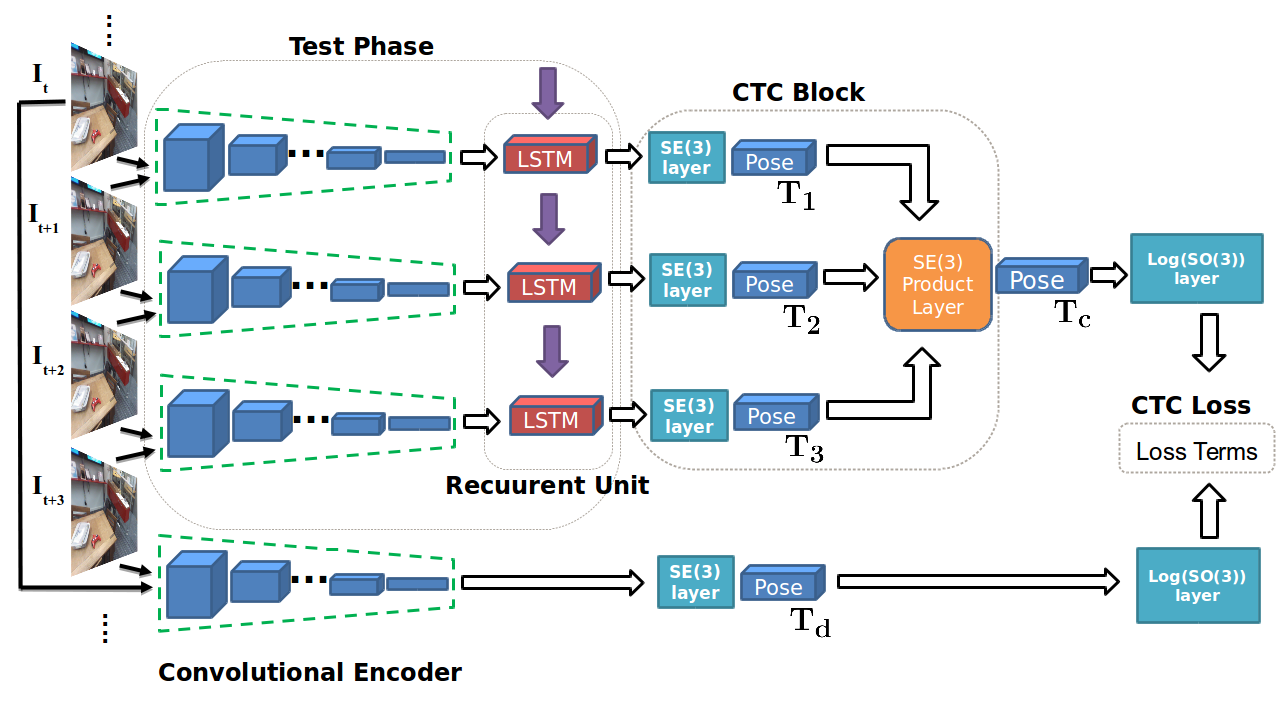}
\end{center}
	\caption{\textbf{End-to-end architecture:} An example of Composite Transformation Constraints (CTCs) being applied to $4$ successive input images. During training, two estimates are generated from the inputs: one for a sequential pairwise constraint and one for a CTC constraint. At test time, each frame is only fed into the network once to receive the output pose from the $SE(3)$ layer. Therefore, the system can run in an online fashion and in real-time on a GPU. (In practice, when training, we use up to $18$ frames in an input window and hence generate multiple CTCs that are applied to frames in the window. Here we show only one CTC block to avoid clutter.)}
\label{fig:long}
\vspace{-0.3cm}
\end{figure*}

Deep learning solutions for VO are a relatively recent but quickly evolving subset of methods for estimating camera ego-motion. While initial approaches relied on ground truth poses for training, recent approaches also explore the possibility of unsupervised training schemes.

\subsection{Supervised Approaches}

Numerous approaches \cite{vinet,L-VO,demon,Wang_IJRR,deepvo} learn the task of VO estimation using ground-truth data available in the form of global-camera poses, recorded by high-precision GPU+IMS rigs.

Konda \textit{et.al.}~\cite{L-VO} first proposed an autoencoder to learn a latent representation of the optical flow between camera frames jointly with the ego-motion estimation task. Kendall \textit{et.al.}~\cite{posenet} proposed a convolutional network based on the GoogLeNet architecture for monocular camera relocalization. Wang \textit{et.al.}~\cite{deepvo} further extend the idea to exploit long term dependencies between monocular frames through a recurrent convolutional network.

Clark \textit{et.al.}~\cite{vinet}, assimilate pose information over windows of sequential frames and their corresponding inertial information using an $SE(3)$ concatenation layer and separately fuse visual and inertial streams to provide robust trajectory estimates. Ummenhofer \textit{et.al.}~\cite{demon} propose 'DeMoN' for supervised joint estimation of depth, ego-motion, surface normals and optical flow given two successive views. They show that learning these multiple-tasks jointly leads to better performance on each of the tasks compared to scenarios where each task was learnt in a disjoint fashion.

Peretroukhin \textit{et.al.}~\cite{dpcnet} recently propose a different approach to supervised VO. Rather than predicting relative transformations between pairs of frames, they train a CNN that \emph{corrects} estimates from an existing VO framework. They use stereo pairs for training and rely on pose graph relaxation to correct existing pose values obtained from SVO~\cite{SVO}. 

However, the training of these networks is supervised against ground truth and is therefore limited by the availability of such recorded ground truth information.

\subsection{Unsupervised Approaches}

Recently, a lot of work has been conducted towards the estimation of depth in a scene, which can be used as a prior to find relative camera pose between associated successive image frames. \textit{Handa et.al.}~\cite{gvnn} in their library gvnn, introduced the 3D spatial transformer. Operating on a depth map along with the corresponding image, it finds the $\se3$ warp $\mbox{\boldmath$\xi$}$ that transforms the camera coordinates of the current frame to those of the next frame, such that when projected back into the image space of the next frame, the photometric error between the resulting $SE(3)$ warped image and the actual next image is minimized. This work paved the way for self-supervised methods that don't require ground truth pose information. Another work along similar lines by Zhou \textit{et.al.}~\cite{sfmlearner} learns both depth and pose from monocular frames, using a novel depth-based pipeline for reconstructing successive frames, although it is unable to recover depth in metric scale. Vijayanarasimhan \textit{et.al.}~\cite{sfm-net} further propose 'SfM-Net' to jointly predict depth, segmentation, optical flow, camera and rigid object motion. They propose both unsupervised and supervised variants based on the availability of ground truth ego-motion or depth.

In a more recent work, Li \textit{et. al.}~\cite{undeepvo_icra} use stereo and monocular geometric constraints to create a composite loss function during training and use only monocular frames for testing.  In contrast, we use an LSTM based architecture that exploits mutliple-views along with their associated pose consistency constraints, while still using frames from a single camera.  

Furthermore, all these unsupervised approaches use depth prediction as a convenient auxiliary task to aid in learning. Our approach is orthogonal to these, in the sense that we rely purely on geometric consistency and do not need such auxiliary tasks for unsupervised learning of VO.




\section{Learning VO without ground-truth labels}

The central idea of this paper is to leverage geometric consistency and use it as a proxy for ground-truth labels. In this section, we describe composite transformation constraints in detail and present our network architecture, loss function, and training details. We also briefly describe how covariance recovery can be easily incorporated into the proposed approach, without additional training overhead.

\subsection{Composite Transformation Constraints}
\label{subsection:CTC}

Composite transformation constraints are based on the fundamental law of composition of rigid-body transformations. Simply put, if we have transformations between two sets of frames $A \mapsto B$ and $B \mapsto C$, then the transform from $A \mapsto C$ is simply the concatenation of the two former transforms. As a toy example (Fig.~\ref{fig:long}), given a sequential set of frames $\mathcal{F} = (I_t, I_{t+1}, I_{t+2}, I_{t+3})$ at time $t$, we train a neural network to predict the transforms: $[T_{t}^{t+1}, T_{t+1}^{t+2}, T_{t+2}^{t+3}]$. Since we do not have access to ground-truth labels, we cannot quantitatively evaluate the accuracy of the predicted transforms. However, for geometrical consistency to hold, we know that the following composite transformation constraints must be satisfied.

\begin{equation}
\begin{aligned}
&  T_{t}^{t+1} \cdot T_{t+1}^{t+2} \cdot T_{t+2}^{t+3} = T_{t}^{t+3}\\
& T_{t}^{t+1} \cdot T_{t+1}^{t+2} = T_{t}^{t+2}\\
&  T_{t+1}^{t+2} \cdot T_{t+2}^{t+3} = T_{t+1}^{t+3}\\
\end{aligned}
\end{equation} 

The extent to which the above constraints are satisfied is a measure of trajectory consistency. We have a convolutional encoder that feeds into a recurrent neural network as our deep architecture for VO estimation (details in Sec \ref{sec:network_architecture}, see Fig. \ref{fig:long}). We first feed all frames in $\mathcal{F}$ into this network and estimate all successive transformations of the form $T_i^{i+1}$. This provides us with all the information required to evaluate the left-hand sides of the above constraints. To evaluate the right-hand sides, we estimate all $T^i_j$s $(j \neq i)$ using only the convolutional encoder and feeding it frames $I_i$ and $I_j$.

As an example, for an image pair $(I_t, I_{t+2})$, the predicted transform $T_t^{t+2}$  must be equal to the product of transforms $T_t^{t+1}$ and $T_{t+1}^{t+2}$, predicted sequentially for frames $(I_t, I_{t+1}, I_{t+2})$. For larger input sequences, we can naturally formulate many more such CTCs. All of them are jointly optimized during the training phase.

Note that, although traditional LSTM-based architectures (without CTCs) would suffice to provide smooth trajectories by mitigating noise between intermediate transforms (smooths them out so that they do not deviate much from the neighboring odometry estimates), it does not ensure geometric consistency of the obtained estimates. The composite transformation constraint is, therefore, essential in bringing about consistency in the predicted sequential transforms, such that the LSTM not only provides smooth trajectory estimates but also estimates that are consistent within the underlying geometry of the trajectory.


\vspace{-0.1cm}
\subsection{Network Architecture}
\label{sec:network_architecture}
\vspace{-0.1cm}

Our network consists of three major components - a convolutional encoder, a recurrent unit, and a CTC block. Fig.~\ref{fig:long} illustrates the proposed end-to-end architecture for unsupervised VO.
\vspace{-0.3cm}

\subsubsection{Convolutional encoder}
Our network follows a similar structure to FlowNetSimple and VGG-11 \cite{Flownet, vgg}. The network takes as input a pair of RGB images, denoted $I_t$ and $I_{t+1}$, stacked along their color channels. We initialize our convolutional layers with the pre-trained weights from VGG-11\footnote{We use a slightly different variant from the one in the original paper \cite{vgg}. Our variants use BatchNorm \cite{batchnorm} before every nonlinearity.}. Unlike VGG-11, our input consists of two images stacked together as opposed to a single image, we replicate and concatenate the weight tensors from VGG-11 to initialize our first layer. The use of pre-trained weights prevents (re-)learning relevant features from scratch. After this initial series of convolutions and pooling, we further aggregate our features globally by using a series of strided fully convolutional layers. During the training process, we continue fine-tuning our weights for the task of estimating $\se3$ transformation parameters. 

The output of our network is a $C-$dimensional vector, $\mathcal{V}$. This vector $\mathcal{V}$ is provided as input to a fully connected layer that regresses a $6$-vector comprising transformation parameters $\xi = (v^T, \omega^T)^T \in \se3$ where $v$ is the translational velocity, and $\omega$ is the rotational velocity respectively.

\begin{figure}[!t]
\begin{center}
   \includegraphics[width=1.0\linewidth]{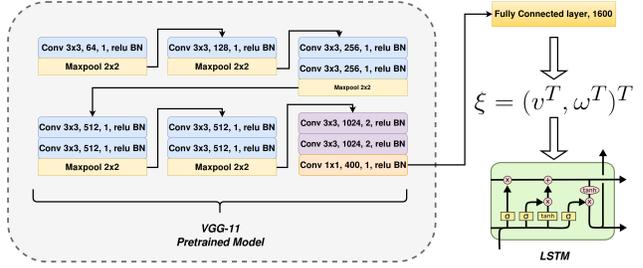}
\end{center}
   \caption{\textbf{Architectural specifics}: Our network builds on the popular VGG-11 network and takes images  $I_{t}$ and $I_{t+1}$ that are resized to $320$x$240$ and then stacked along the RGB channels. Each convolution layer is followed by a ReLU non-linearity. Then, batch normalization and max-pooling are successively applied. Finally $2$ layers of strided convolution are applied followed by a $1$x$1$ convolution layer to produce a latent vector of length $1080$ that is used as an input to the LSTM unit. The LSTM has $1000$ units in its hidden state. A final fully-connected layer maps the output of the LSTM to a $6$-dimensional $\se3$ coordinate vector.}
\label{fig:short}
\label{fig:threecol}
\vspace{0.5cm}
\end{figure}

\subsubsection{Recurrent unit}
The vector $\mathcal{V}$ from the convolutional encoder is also input to a recurrent unit that maintains a hidden state. It regresses the $\se3$ transformation parameters for each frame in the input sequence. We use LSTM units as recurrent units throughout this paper. We discuss the training methodology in further detail in Sec.~\ref{sec:training_details}.
\vspace{-0.25cm}

\subsubsection{CTC block}
The CTC block gathers outputs from the encoder and the recurrent units and applies CTCs to them. It is built using various layers that perform lie-algebraic operations. We briefly describe each layer in this block.

\textbf{\textit{SE(3) Layers}}: The ${SE(3)}$ layers are responsible for mapping the estimated $\se3$ parameters into the corresponding $SE(3)$ transformation matrix and vice versa. A 3D rigid body transformation $\mathbf{T}\in SE(3)$ is a rotation $R$ and translation $t$ in 3D space, and is defined as follows.
\begin{equation}
   T = \left( \begin{array}{cc} R & t \\ 0 & 1 \end{array}\right) \text{where } R \in SO(3) \thinspace \text{and } t \in \mathbb{R}^3
\end{equation}

We denote a local transformation between the camera poses at times $t$ and $t+1$ as ${T_{t}^{t+1}}$. We use $\xi = (v^T \enspace \mathfrak{\omega}^T) \in \se3$ to define the lie-algebraic exponential coordinates of the local transformation.

An element in $\se3$ can be mapped to one in $SE(3)$ by using the exponential map, which is simply the matrix exponential over a linear combination of the generators of the tangent-space at the identity element of the Lie group. The exponential map can be inverted to obtain the logarithm map from $SE(3)$ to $\se3$. Our implementations of the SE(3) layers use the exponential and logarithm maps and their corresponding small-angle approximations presented in \cite{lie_eade}.

\textbf{\textit{CTC computation}:} The network has multiple CTC blocks, where each such block is responsible for the computation of one particular composition constraint. In effect, a CTC block computes the constraint in the following manner. It first obtains $\se3$ estimates for the left-hand side (LHS) of the constraint from the recurrent block, maps them to transformation matrices in $SE(3)$, and composes (concatenates) all of them. Then, it obtains an independent $\se3$ estimate for the right-hand side (RHS) of the constratint. The LHS and RHS estimates are then passed to the $\se3$ loss layer described below, which evaluates the constraint and computes gradients.





\subsection{Loss Functions}

Our complete loss function consists of a CTC error term and a regularization term. Our loss terms are generic, and can be applied to supervised as well as unsupervised settings. 

\textbf{\textit{CTC Loss:}} This loss is dictated by the composite transformation constraints as described in Sec~\ref{subsection:CTC}. It is computed between a direct transformation $T_d$ predicted between non-consecutive frames and a composite transformation $T_c$, composed as a product of smaller sequential transformations resulting from the predictions for successive frames:

\begin{equation}
    \mathcal{L}_{ctc} = { \| \xi_d - \xi_c \|}^2_2 
\end{equation}
\noindent
where $\xi_d, \xi_c$ are the $\se3$ exponential coordinates for the transforms $T_c$ and $T_d$, respectively.

\textbf{\textit{Regularization Term:}} 
While the unsupervised term $\mathcal{L}_\text{ctc}$ is useful for enforcing consistency, using the above term alone could result in a degenerate solution where the network can predict zeros for $\xi_c, \xi_d$. To prevent this collapse to a trivial solution, we introduce a regularization loss term. Specifically, we assume a prior on each of the transforms estimated by the network from a standard VO pipeline. Such a prior aids in avoiding trivial solutions and is inexpensive to obtain. For each transform $\xi_*$ predicted by either the convolutional encoder or the recurrent unit, we have a prior $\hat{\xi}_*$ from a conventional visual odometry estimator, used as a regularization term.
\begin{equation}
    \mathcal{L}_{reg} = { \| \xi_* - \hat{\xi}_* \|}^2_2
\end{equation}
Again, it's essential to note that this estimator can be very noisy, and is used only in a \emph{supporting role} to the CTC loss term.

The overall loss function is a weighted sum of the CTC loss and the regularization term. In the expression below, $\alpha, \beta$ are scalar weights associated with each of the loss terms.

\begin{equation}
    \mathcal{L}_{final} = \alpha \mathcal{L}_{ctc} + \beta \mathcal{L}_{reg} 
\end{equation}

\subsection{Covariance recovery}
In VO, recovering the covariance of an estimate is very important, as it can be efficiently exploited when recovering global information using pose-graph optimization (e.g., in \cite{orb2}). Kendall~\textit{et al.} \cite{bayesianPosenet} use dropout \cite{dropout} as a means of bayesian approximation to recover covariance from relocalization estimates from a trained CNN.

Similarly, we use dropout at the penultimate fully connected layer of the convolutional encoder as well as the recurrent blocks. At test time, instead of removing dropout from the network, we retain dropout layers and generate $K$ predictions for each input pair of frames. While generating each of these $K$ estimates, we randomly drop a fraction $\gamma$ of the units of the penultimate fully connected layer, which results in a different estimate in each pass. The hypothesis is that, if the network is very \emph{confident} of its estimates, then the variance in the obtained samples must be low. We fit a Gaussian density function to the $K$ samples and use this density for covariance recovery.

\subsection{Training Details}
\label{sec:training_details}
For ease and flexibility during training, we divide our training process into two stages: a pre-training phase and a sequential training phase:

\textbf{\textit{Pretraining phase:}} While we can train in an end-to-end fashion, we consider the option of pretraining the convolutional layers in order to provide structured  and informative latent features as input to the LSTM during sequential training. 
The pre-training phase consists of training the output of the convolutional encoder against \emph{noisy} VO estimates from a traditional odometry estimation framework. In the case of unsupervised training, when ground truth data may not be available, we rely on the frame-to-frame transformation estimates provided by RGB-D ORB-SLAM \cite{orb2}. While these estimates are noisy, they provide a fair starting point for the network to learn from.

\textbf{\textit{Sequential training phase:}} The sequential training phase consists of providing the network \emph{windows} (sequences) of frame pairs as input. 

\begin{figure*}[!th]
\begin{center}
    \includegraphics[width=0.75\textwidth]{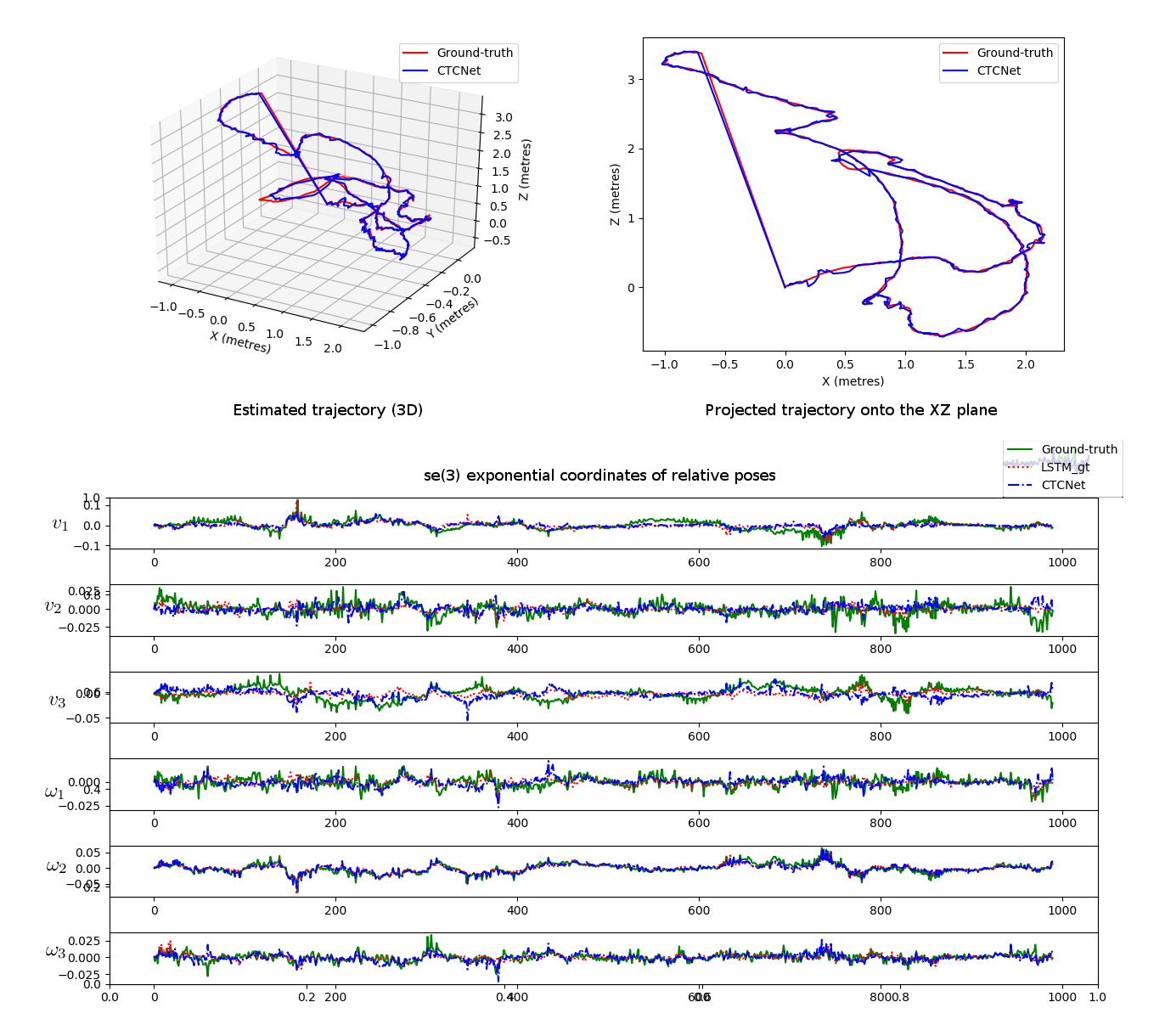}
\end{center}
    \caption{\textbf{Trajectory estimates} on a sequence from the 7-Scenes test split. Top (left to right): Output trajectories are shown in \textcolor{red}{red}, against ground-truth trajectories in \textcolor{blue}{blue}. Bottom: $\se3$ estimates of relative poses. Each of the $6$ $\se3$ coordinates is plotted independently. On this sequence, CTCNet performs better than LSTM$_{gt}$.}
\label{fig:trajAndError}
\vspace{-0.3cm}
\end{figure*}

For training the network we use the Adam optimizer \cite{ADAM}, with an initial learning rate of $10^{-4}$, and momentum equal to $0.9$. We decrease the learning rate by a factor $0.5$ every few epochs. We train for a total of $40$ epochs. During the pre-training phase, we only use the regularization loss term. During sequential training, we start with sequences of length 3 frame pairs, i.e. [$I_t$..$I_{t+3}$], and gradually increase to 18 frame pairs. We use several composite transformation constraints for each window, as well as regularization terms, with initial coefficient values $\alpha = 1$ and $\beta = 3$. We also experiment with end-to-end training of the full network, using the same loss terms. To prevent over-fitting, we  apply a dropout of $0.7$ at the penultimate fully connected layer of the convolutional encoder and the recurrent unit.

\vspace{-0.15cm}
\section{Experiments and Results}
\vspace{-0.15cm}

In this section, we describe the experiments carried out to analyze the efficacy of the proposed approach, and the findings we made in the process. We begin by describing the basic setup for various experiments, and then describe several variants of deep architectures that were evaluated. We then present qualitative, as well as quantitative comparisons and proceed to a discussion of further scope for work.
\vspace{-0.25cm}

\begin{table*}[!h]

\label{table:results_main}
\centering
\resizebox{\textwidth}{!}{
\begin{tabular}{|c|c|c|c||c|c|c|}
\hline
\textbf{Network Architecture} & \multicolumn{3}{c||}{\textbf{Absolute Trajectory Error (ATE) (meters)}} & \multicolumn{3}{c|}{\textbf{$\se3$ error (L2-distance)}} \\
\hline
 & \textbf{\textit{redkitchen}-03} & \textbf{\textit{office}-02} & \textbf{\textit{fire}-04} & \textbf{\textit{redkitchen}-03} & \textbf{\textit{office}-02} & \textbf{\textit{fire}-04}  \\
\hline
$\textnormal{CNN}_{gt}$ & $0.0274 \pm 0.0080$ & $0.1216 \pm 0.0820$ & $0.1421 \pm 0.0436$ & $0.0293$ & $0.0364$ & $0.0354$ \\
$\textnormal{LSTM}_{gt}$ & $0.0242 \pm 0.0091$  & $0.1119 \pm 0.0327$ & $0.1101 \pm 0.0412$ & $0.0257$ & $0.0253$ & $0.0291$ \\
$\textnormal{CNN}_{unsup}$ & $0.0402 \pm 0.0121$  & $0.1394 \pm 0.1027$ & $0.1707 \pm 0.0647$ & $0.0382$ & $0.0401$ & $0.0368$ \\
$\textnormal{LSTM}_{unsup}$ & $0.0392 \pm 0.0121$  & $0.1290 \pm 0.0670$ & $0.1700 \pm 0.0513$ & $0.0343$ & $0.0400$ & $0.0359$ \\
$\textnormal{CNN}_{aug}$ & $0.0787 \pm 0.0562$  & $0.1662 \pm 0.0908$ & $0.1675 \pm 0.0833$ & $0.0605$ & $0.0772$ & $0.0547$ \\
$\textnormal{LSTM}_{aug}$ & $0.0780 \pm 0.0531$  & $0.1318 \pm 0.0613$ & $0.1486 \pm 0.0809$ & $0.0352$ & $0.0395$ & $0.0345$ \\
\hline
$\textnormal{ORB-SLAM}$ & $0.0326 \pm 0.0140$  & $0.1005 \pm 0.0620$ & $0.1057 \pm 0.0515$ & $0.0352$ & $0.0426$ & $0.0305$ \\
\hline
\hline
$\textbf{CTCNet}$ & $0.036\pm0.0012$ & $0.1226\pm0.0183$ & $0.12918\pm0.0246$ & $0.0286$ & $0.0384$ & $0.0338$ \\
\hline
\end{tabular}
} 
\caption{\textbf{Ablation analysis} of the proposed network architecture and variants. We evaluate the absolute trajectory error (ATE) (in meters). We also evaluate the relative pose estimation error in $\se3$ exponential coordinates ($L2$-distance).}
\vspace{-0.3cm}
\end{table*}

\subsection{Dataset and Metrics}
\vspace{-0.15cm}
\subsubsection*{Dataset} 
\vspace{-0.15cm}
Although most approaches \cite{vinet,L-VO,undeepvo_icra,dpcnet,sfm-net,Wang_IJRR,deepvo,sfmlearner} evaluate their approach on the KITTI \cite{Kitti} benchmark, the camera in KITTI moves on the road plane and does not exhibit unconstrained $6$ DoF motion.  Wang \emph{et. al.} \cite{Wang_IJRR} present results on a wide range of datasets, but their approach notably performs poorly when camera motion is unconstrained. To provide baseline results for several deep architecture on a challenging dataset, we conduct our experiments on the Microsoft 7-Scenes \cite{7scenes} dataset, which is increasingly being used to evaluate VO and/or relocalization performance \cite{posenet,auxlearning}. The dataset consists of tracked $640 \times 480$ resolution RGB camera frames collected using a handheld Microsoft Kinect camera. Although depth information is available, only the RGB images are used as input to all network variants we consider during training as well as testing.

The dataset consists of $7$ scenes, with a total of $46$ sequences comprising of about $1000$ frames each. We use the dataset-provided train/test splits for all our experiments. During the initial training phase, we often use frames that are randomly separated between $1$-$5$ time steps apart in the same sequence. This allows for a wider range of transformations and allows for a higher number of training pairs. Overall, we composed a total of $49152$ image pairs for training, $25686$ image pairs for validation, and $15983$ image pairs (17 sequences) for testing.

\subsection*{Metrics}
\vspace{-0.15cm}
To compare the output trajectories of our approach with ground-truth, we use the absolute translation error (ATE) metric. Further, to evaluate the accuracy of relative pose estimation, we also analyze the $L2$-distance between the estimated $\se3$ exponential coordinates and the corresponding ground-truth $\se3$ vector.



\subsection{Network Architectures Evaluated}
\vspace{-0.15cm}

We carry out extensive experiments on several variants of deep network architectures (supervised, unsupervised, stateless, stateful) for VO prediction, to analyze the benefits and pitfalls offered by each. Here, we enumerate each of the variants tested. The supervised variants are provided ground-truth pose estimates for supervision, whereas the unsupervised variants are trained without ground-truth pose information.
\vspace{-0.2cm}
\begin{itemize}
\item $\textnormal{CNN}_{gt}$: The convolutional encoder supervised using ground-truth pose information. \vspace{-0.1cm}
\item $\textnormal{LSTM}_{gt}$: The convolutional encoder with its output fed to the recurrent unit. \vspace{-0.1cm}
\item $\textnormal{CNN}_{unsup}$, $\textnormal{LSTM}_{unsup}$: Unsupervised variants of $\textnormal{CNN}_{gt}$ and $\textnormal{LSTM}_{gt}$, respectively. \vspace{-0.1cm}
\item $\textnormal{CNN}_{aug}$, $\textnormal{LSTM}_{aug}$: Similar to $\textnormal{CNN}_{gt}$, and $\textnormal{LSTM}_{gt}$ respectively. However, instead of simply taking in odometry estimates from ORB-SLAM \cite{orb2}, every time an image pair and its corresponding odometry esimate are drawn from a Gaussian distribution centered about the ORB-SLAM estimate, to account for noisy estimates. \vspace{-0.1cm}
\item \textbf{CTCNet}: The proposed architecture that enforces composite transformation constraints (CTCs). 
\vspace{-0.2cm}
\end{itemize}


\begin{figure}[!b]
\centering
\includegraphics[width=0.5\textwidth]{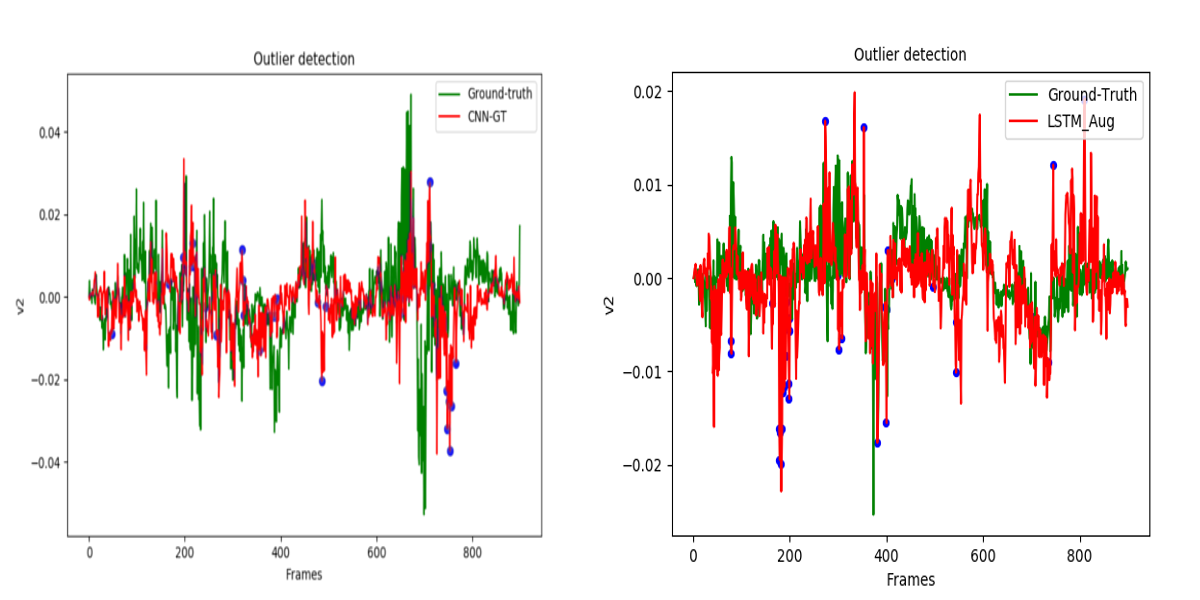}
\caption{\textbf{Outlier detection}: Upon covariance recovery, estimates with covariance above a threshold are marked outliers (here shown in \textcolor{blue}{blue}).}
\label{fig:outlier}
\end{figure}

\subsection{Results}

\begin{figure*}[!h]
\centering
\includegraphics[scale=0.8]{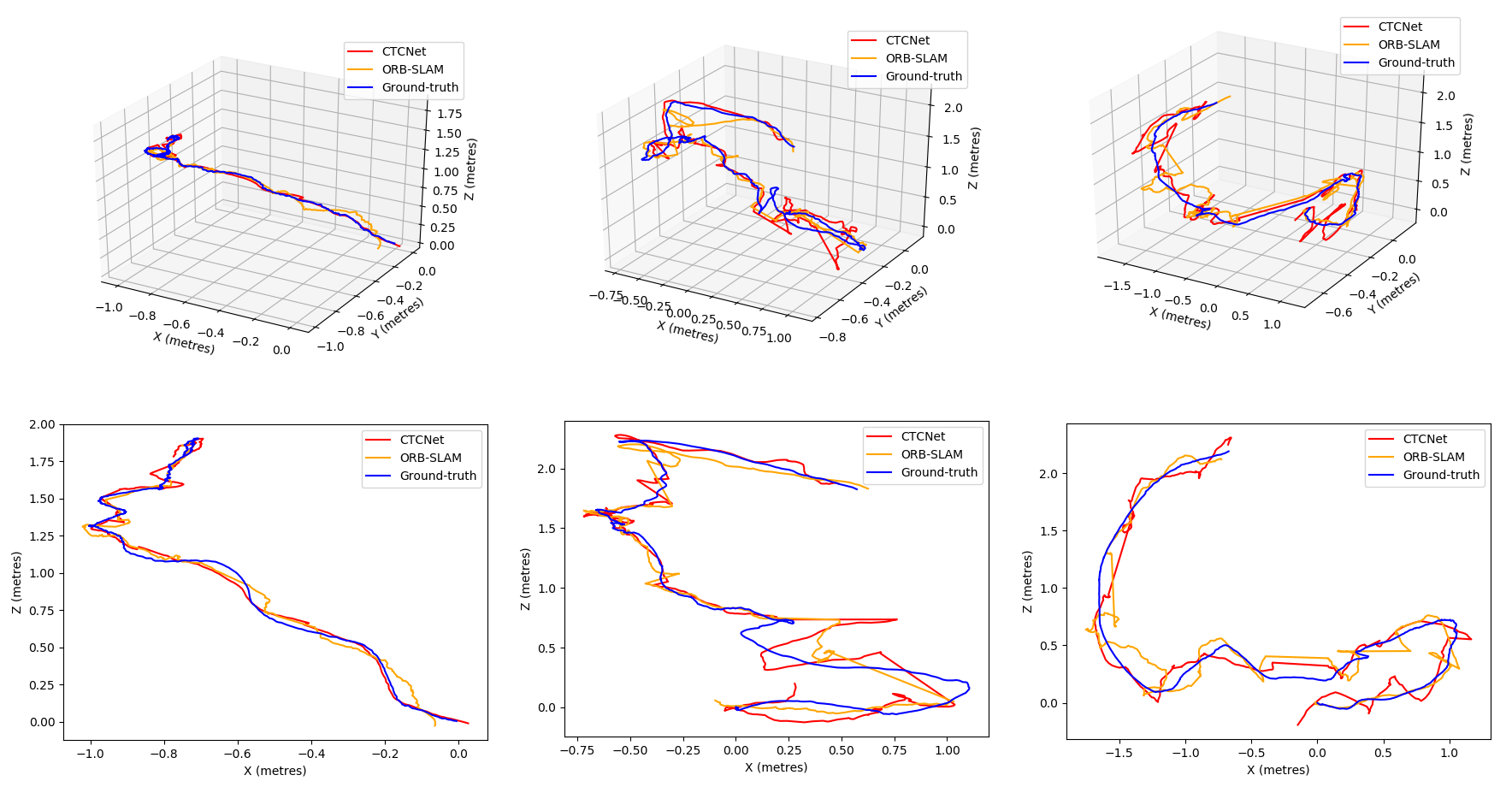}
\caption{\textbf{Estimated trajectories}: (Top) Estimated 3D trajectories from CTCNet and ORB-SLAM plotted againscat ground-truth. (Bottom) 2D projections of these trajectories onto the XZ-plane.}
\label{fig:trajectories}
\vspace{-0.4cm}
\end{figure*}

We evaluate all network variants on the test split of the 7-Scenes dataset and the analysis is presented in Table \ref{table:results_main}. As one would expect, the networks trained against ground-truth turned out to be the best-performing models. Understandably, the LSTM variants achieved better performance compared to their convolutional counterparts, due to the additional context they store in their hidden state.

It can, however, be seen that CTCNet performs on par with supervised approaches, although it has been trained only using \emph{noisy} estimates from a VO pipeline. Moreover, LSTM variants that were trained purely against the noisy estimates (i.e., without using CTC) perform poorly. Data augmentation / label noise shows a slight improvement in ATE, as evident from the CNN$_{aug}$ and LSTM$_{aug}$ results. In certain sequences, CTCNet achieves significantly lower error compared to CNN$_{gt}$ and LSTM$_{gt}$ (Fig.~\ref{fig:trajAndError}). This makes a strong case for using geometric consistency for unsupervised learning, especially in tasks such as visual odometry and SLAM.
\vspace{-0.2cm}
\subsubsection{Comparision with ORB-SLAM}
\vspace{-0.15cm}
From Table~\ref{table:results_main}, we observe that CTCNet does better in terms of relative pose estimation  when compared to ORB-SLAM and hence has a lower $\se3$ error ($L2$-distance metric). However, ORB-SLAM does marginally better on ATE. This suggests that CTCNet performs better \emph{locally}, whereas ORB-SLAM is better at a \emph{global} level. Since ORB-SLAM is using keyframes it is able to optimize over an entire sequence of images with a similar viewpoint no matter how long it is. We believe this can be mitigated by training CTCNet on longer window lengths (currently it takes in a window of only $18$ image pairs as input), but being able to flexibly control this window size the way that model-based approaches are able to\footnote{ORB-SLAM has this flexibility built-in, using keyframes. At a frame rate of 30 fps, a window-size of $18$ frames would mean that ORB-SLAM has a very stable keyframe that boosts performance.} is not currently addressed in deep VO approaches and forms an interesting avenue for future work.   



\vspace{-0.15cm}
\subsubsection{Uncertainty estimation}
\vspace{-0.15cm}

Fig.~\ref{fig:outlier} illustrates the performance of the proposed covariance recovery scheme. Using a dropout (with drop ratio $10\%$), we draw $10$ estimates per input pair using the CNN$_{gt}$ model. We plot the estimated relative $\se3$ coordinates with respect to those from ground-truth transforms. Fig.~\ref{fig:outlier} shows this plot along the dimension $v_2$, i.e., translational velocity along the Y-axis. If the covariance of a particular estimate is very high (i.e., if the $10$ estimates drawn have their variance above a set threshold), that estimate is characterized as an \emph{outlier} (shown in blue). We see that the network reasonably detects and characterizes several outliers. This piece of information is valuable, especially in weighing these estimates when constructing a global representation (using pose graph optimization, for instance). Moreover, this covariance recovery need not be learned. It suffices if dropout layers are present during training. 
Investigation of how the estimated uncertainty can be exploited to suppress the effect of outliers (see Fig.~\ref{fig:trajectories}) is deferred to future work.

\vspace{-0.35cm}
\subsubsection{Generalization}
\begin{figure}[!ht]
\vspace{-0.35cm}
\centering
\includegraphics[width=0.45\textwidth]{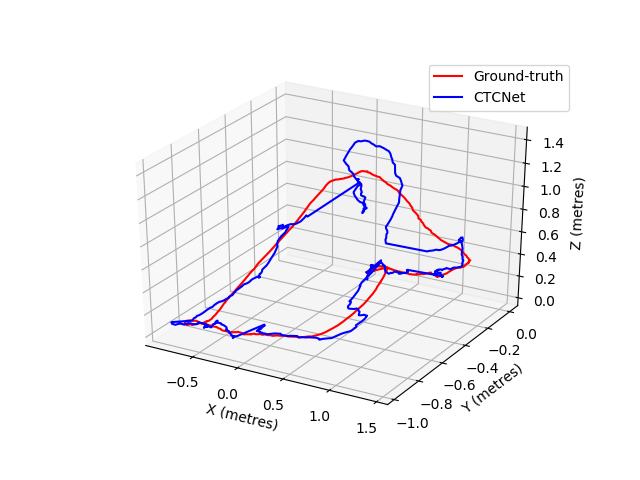}
\caption{\textbf{Generalization to unseen data}: CTCNet was evaluated on a sequence that was in stark contrast to the kind of sequences it had been presented with during training. Estimated 3D trajectory plotted against ground-truth.}
\label{fig:generalization}
\vspace{-0.6cm}
\end{figure}

\vspace{-0.15cm}
We also evaluate CTCNet in scenarios that it has never encountered during training. To do so, we train CTCNet using only $4$ scenes from the 7-Scenes dataset (\textit{chess}, \textit{office}, \textit{redkitchen}, and \textit{stairs}). We evaluate VO estimation performance on a sequence from the \textit{fire} scene and report the obtained trajectory in Fig.~\ref{fig:generalization}. This sequence bears no resemblance to the training data presented to CTCNet, either during the pre-training phase or the sequential training phase. However, it has been trained on very little data ($4$ scenes); training on more should improve performance. Moreover, CTCNet alleviates the need for hyperparameter tuning, which is frequently required for traditional VO pipelines such as ORB-SLAM \cite{orb2}.

\section{Conclusion}
\vspace{-0.15cm}
In this paper, we showcase a new end-to-end architecture for self-supervised training to regress pose transformations between monocular frame-pair sequences. We demonstrate the use of a differentiable flexible composite constraint and its application in both supervised and unsupervised settings. Our method works well in the supervised setting with reduced ATE, when tested on indoor sequences. In the future, we plan to extend the work to a full-fledged SLAM system. We would also look to generate and utilize depth information (RGB-D) of sequential scenes for dense reconstruction and trajectory estimation.

{\small
\bibliographystyle{ieee}
\bibliography{references}
}

\end{document}